\documentclass[letterpaper, 10 pt, conference]{ieeeconf}  

\IEEEoverridecommandlockouts                              

\title{\LARGE \bf
Joint Machine-Transporter Scheduling for Multistage Jobs with Adjustable Computation Time
}

\author{Koresh Khateri and Giovanni Beltrame
\thanks{This work was supported by MITACS and Spiri Robotics}
\thanks{The authors are with Department of Computer and Software Engineering, Polytechnique Montréal, 2500 Chem. de Polytechnique.
        {\tt\small khateri.koresh@polymtl.ca} and {\tt\small giovanni.beltrame@polymtl.ca}}%
}

\usepackage[dvipsnames]{xcolor}
\usepackage[acronym]{glossaries}
\usepackage{hyperref}
\makeglossaries
\newacronym{ffspt}{FFSPT}{Flexible Flow Shop Problem with Transport}
\newacronym{ffsp}{FFSP}{Flexible Flow Shop Problem}
\newacronym{fsp}{FSP}{Flow Shop Problem}
\newacronym{fjsp}{FJSP}{Flexible Job Shop Problem}
\newacronym{jsp}{JSP}{ Job Shop Problem}
\usepackage{arrayjobx}
\usepackage{tikz}
\usetikzlibrary{arrows.meta}
\usetikzlibrary{positioning}
\usetikzlibrary{calc}
\usepackage{bbm}
\usepackage{color}
\usepackage{amsmath}
\newtheorem{remark}{Remark}

\usepackage[caption=false, font=footnotesize]{subfig}

\usepackage{graphicx}
\usepackage[ruled,vlined]{algorithm2e}
\usepackage{todonotes}
\usepackage{placeins}
\usepackage{optidef}
\usepackage{url}

\SetKwProg{Fn}{Function}{}{}
\SetKwRepeat{Repeat}{repeat}{until}
\SetKwProg{Fn}{Function}{}{}
\begin{document}
\LinesNumbered

\maketitle

\begin{abstract}
  This paper presents a scalable solution with adjustable computation time for the joint
  problem of scheduling and assigning machines and transporters for missions that
  must be completed in a fixed order of operations across multiple stages. A battery-operated multi-robot system with a maximum travel range is employed as the transporter between stages and charging them is considered as an operation. Robots
  are assigned to a single job until its completion. Additionally, The operation completion time is assumed to be dependent on the
  machine and the type of operation, but independent of the job. This work aims to minimize
  a weighted multi-objective goal that includes both the required time and energy
  consumed by the transporters. This problem is a variation of the
  flexible flow shop with transports, that is proven to be NP-complete. To
  provide a solution, time is discretized, the solution space is divided temporally,
  and jobs are clustered into diverse groups. Finally, an integer linear
  programming solver is applied within a sliding time window to determine assignments and
  create a schedule that minimizes the objective. The computation time can be
  reduced depending on the number of jobs
  selected at each segment, with a trade-off on optimality. The proposed
  algorithm finds its application in a water sampling project, where water sampling jobs are assigned to robots, sample deliveries at laboratories are scheduled, and the robots are routed to charging stations.
\end{abstract}

\IEEEpeerreviewmaketitle

\section{Introduction}
Scheduling and machine assignment for optimizing time and energy has been
extensively analyzed in different
settings~\cite{ruiz2010hybrid,lee2019review,xiong2022survey}. This paper considers a
water sampling scenario in which robots pick up water samples, deliver them to
a set of laboratories, and recharge themselves in a dedicated facility. The robots'
constraints are limited travel range, the capacity of chargers, laboratories,
and base stations. The goal is to provide an optimal or close to optimal
assignment and scheduling of robots, laboratories, chargers, and base stations.
this paper provides an optimization heuristic for a generalized form of this problem
where multi-staged jobs with a similar sequence of operations are scheduled and
assigned to the machines and transporters with a limited travel range.

Similar problems have been considered in
industry~\cite{cunha2020deep,ibrahim2023improved} focusing on machines'
processing time and constraints to optimize machine utilization in factories or
job shops. The expansion of the manufacturing plants and the rise of distributed
manufacturing necessitated considering transportation time. Several authors have
studied the scheduling of transporters alongside machine
sequencing~\cite{naderi2009improved} considering travel time as a key factor and
assuming an unlimited travel range and number of transporters. However, with the
introduction of robots as transporters, other limitations like the travel range
and availability cannot be ignored~\cite{Liang2022Integrated}, but the
complexity of the problem increases further when coupled with transporter
scheduling. Soukhal et al.~\cite{SOUKHAL200532} prove that a two-machine
\acrlong{ffspt} (\acrshort{ffspt}) is strongly NP-hard.
Several studies have attempted to address the combined
problem~\cite{GHEISARIHA2021152,ren2022joint, wang2022flow}, using exact,
heuristic, and meta-heuristic algorithms. However, these studies consider
small-scale scenarios and due to the inherent complexity of NP-hard
combinatorial problems, these solutions are not scalable. In this paper, we
propose a specific class of joint scheduling and assignment of machines and
transporters (as in the water sampling application) in which the problem is
dividable in the time dimension and it clusters jobs based on their properties to
be able to optimize a large-scale problem.

\section{Related Work}

The \acrlong{jsp} (\acrshort{jsp}) considers scheduling $N$
  jobs that each must be performed in its given order. The machines can only perform
  one operation at a time, and there is a processing time associated with
  completing an operation on a machine~\cite{JAIN1999390}. A
  variant of \acrshort{jsp} is the \acrlong{fsp} (\acrshort{fsp}) in which there is an identical
  flow of operations for each job (i.e., the process route is identical) and the
  $s$-th operation of a job must be performed on the $s$-th machine.

  The \acrshort{ffsp} is a generalized version of \acrshort{fsp} where parallel
  machines are arranged in multiple stages and there is at least one stage
  containing two or more machines that are able to perform the same
  operation~\cite{wang2005flexible}.
  Recently, several studies are concerned with using automated robots or
  human-operated transporters to transport between different
  machines~\cite{AMIRTEIMOORI2022108672,GHEISARIHA2021152,naderi2009improved}
  that is called
  \acrshort{ffspt}~\cite{AMIRTEIMOORI2022108672,GHEISARIHA2021152,SOUKHAL200532,wang2022flow}.
  This paper's problem is a specific class of \acrshort{ffspt} with range
  limited transporters where the transporters are assigned to a job until it is
  finished, the problem is defined mathematically in Section~\ref{sec:problem}.
  The \acrshort{ffsp} formulation has been used in the scheduling and assignment
  of several projects. It is encountered in multiple real-world applications and
  is used in the process optimization of several industries like steel
  production~\cite{long2018scheduling}, producing LCDs~\cite{choi2011real}, and
  concrete production~\cite{rodriguez2005performance}. To the best of our
  knowledge, there is no work applying the \acrshort{ffspt} formulation to a
  multi-robot sample collection system.

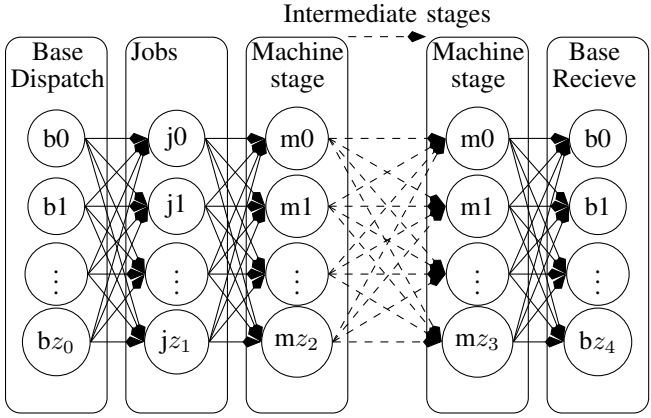
\begin{figure}[h]
  \centering
  \begin{tikzpicture}
    \newarray\machineLbl
    \newarray\zLbl
    \readarray{machineLbl}{b&j&m&m&b}
    \readarray{zLbl}{$z_0$&$z_1$&$z_2$&$z_{S-2}$&$z_{S-1}$}
    \newarray\stageLbl
    \readarray{stageLbl}{{Base\\[-2]Dispatch}&Jobs&{Machine\\[-2]stage}&{Machine\\[-2]stage}&{Base\\[-2]Recieve}}
    \def\offset{{0,0,0,0.8,0.8,0.8}}

    \foreach \i in {1,2,...,5}{ 
      \node[draw, minimum width=1.35cm, minimum height=5cm, rounded corners=2mm] (r\i) at (\i*1.6+\offset[\i-1],0) {};
        \node[anchor=north west, inner sep=2pt, align=center] at (r\i.north west) {\stageLbl(\i)};
        \foreach \j in {1,2,...,4}{
          \pgfmathtruncatemacro{\hi}{\i-1}
          \pgfmathtruncatemacro{\hj}{\j-1}
            \newarray\textCircle
            \readarray{textCircle}{\machineLbl(\i)\hj&\machineLbl(\i)\hj&\vdots&\machineLbl(\i)$z_\hi$}
            \node[draw, circle, anchor=center] (c\i\j) at ($(r\i.north)!0.18*(\j+.5)!(r\i.south)$) {\textCircle(\j)};    
        }   
    }

    \def\arrowType{{"solid","solid","dashed","solid"}}
    \tikzset{bKite/.tip={Kite[length=2.5mm,width=1.5mm]}}
    \foreach \i in {1,2,...,4}{
        \foreach \j in {1,2,...,4}{
            \foreach \k in {1,2,...,4}{
                \pgfmathsetmacro{\usePattern}{\arrowType[\i-1]}
                \draw [-bKite,  \usePattern]  (c\i\j.east) -- (c\the\numexpr\i+1\relax\k.west);
            }
        }
    }
    \draw [-bKite, dashed] (r3.north east) -- (r4.north west) node[pos=0.5, above] {Intermediate stages};
  
  \end{tikzpicture}
  \caption{Jobs with several stages of operations with parallel machines at each stage.}
  \label{fig:graphStages}
  \end{figure}

  The proposed method divides the problem into several subproblems in such a way as to minimize
  the loss of optimality. Exact, heuristic and metaheuristic algorithms have
  been developed for solving this problem~\cite{RUIZ20101}. Exact methods like the branch and bound are only used for toy examples like a two-machine two-stage
  setting~\cite{arthanary1971extension}. Lee \& Kim~\cite{lee2004branch} show
  that the branch and bound can provide a solution in a reasonable time for up to
  15 jobs in a 2-stage problem. Heuristics that are closer to the proposed approach
  involve dispatching rules that are mostly used in dynamic environments where
  jobs are arriving in real-time. Li et al.~\cite{li2013heuristic} provide a new
  heuristic dispatching rule $H'$ and compare it with classical dispatching
  rules. The shortest processing time dispatching rule follows a scheduling
  approach where jobs are arranged in increasing order of priority.
  Specifically, jobs are scheduled starting from the one with the shortest
  processing time and proceeding toward the one with the longest processing
  time.

  Additional research endeavors have also explored variations of
  \acrshort{ffsp}: Defersha~\cite{defersha2015simulated} considers stage
  omissions and several works include tardiness
  minimization~\cite{nejjarou2023inspired,GHORBANISABER2022105604,oujana2021linear}.

\section{Problem Formulation}\label{sec:problem}
  
Consider a set of $N$ jobs $J=\{j_0, j_1, \cdots, j_{N-1}\}$ where each job has
a starting location (e.g., to acquire a sample of water at a certain location,
or to pick up an object from a warehouse) and consists of $S$ operations
$O=\{o_0, o_1, \cdots, o_{S-1}\}$ that have to be performed in order in $S$
stages. Each stage $s$ has $G_s$ parallel machines $M_s=\{m_{0,s}, m_{1,s},
\cdots,m_{G_s-1,s}\}$ that can perform $o_s$ with processing time $pt(m_{k,s})$
and capacity $c(m_{k,s})$ (i.e., the maximum number of operations that can be
performed in parallel by $m_{k,s}$ at any time $t$) and $c(m_{k,s},t)$ as the number of under process operations by $m_{k,s}$. Furthermore, each machine
has a geographical location $x$. We define the \emph{transport time}
$\text{tt}(m_{q,s_h},m_{r,s_{h+1}})$ (i.e., transport time between machine $q$
at stage $h$ and machine $r$ at stage $h+1$) and, similarly, the \emph{transport
  energy} $\text{te}(m_{q,s_h},m_{r,s_{h+1}})$.

We assume that there are transporter \emph{base stations} $b_i \in B$ that contain $A(b_i,t)$ available and charged transporters at time $t$ with a capacity $c(b_i)$ (i.e., the
maximum number of transporters that can be stored at $b_i$ at any time, available or not) and $c(b_i,t)$ as the number of stored transporters at time $t$ in $b_i$. Each
base station has a location and a transporter is \emph{dispatched} from a base
station $b_d$, visits the job location, and then proceeds with different stages
of processing on the machines, finally returning to a \emph{receiving} base
station $b_r$ as shown in Fig~\ref{fig:graphStages}. We assume that once a transporter is assigned to a job $j_i$, it cannot be reassigned until
the completion of $j_i$. Let a facility be a general term referring to a machine or base station

The problem is to assign a schedule $P$ to a sequence of machines and two base
stations (the dispatch and receiving base stations) for each job $j_i$:
\[
  P(j_i)={b_d|t_D,j_i|t_1,m_{b,2}|t_2,\cdots,
    m_{z,S-2}|t_{S-2},b_r|t_R}
\]
where $m_{x,y}|t_z$ is the arrival of the transporter to the machine $m_x$ of stage $y$ at time $t_z$.
Similarly, $b_d|t_D$, $j_i|t_1$, $b_r|t_R$ represents assigning a
transporter from base station $b_d$ where a transporter is dispatched at time
$t_D$ to visit job $j_i$'s initial location at time $t_1$, and return to base
station $b_r$ at time $t_R$. In the following, we use the notation $P,j$ instead
of $P(j_i),j_i$ for brevity when there is no ambiguity. Subscript indexing is used for referring to the timing sequence of $P$ and superscript for the facility sequence (i.e., $P^s$ is the machine at stage $s$ and $P_s$ is the timing to arrive (dispatch time in case of $s=0$) at stage $s$ of $P$) 


A job $j_i$ is finished after the transporter is at its receiving base station
and has been recharged (i.e. processed) at the time
$t_f(P)=P_{S-1}+pt(P^{S-1})$. The energy used for job $i$ is the sum of
the transport energy of each stage $E(P)=\Sigma_{0 \leq s \leq S-2}te(P^s,P^{s+1})$. Type of a facility $\text{type}(P^s)$ is defined to be the type of
$P^s$ (i.e., base dispatch, job, machine, base receive types), and $\text{attached}(P^s)$ is a function that returns 0 if the transporter is not needed while the operation is processing (e.g., delivering a
sample to a laboratory), 1 otherwise. Our goal is to find job schedules that:

\begin{equation}
  \begin{aligned}\label{eq:objective}
  \min_{P} \quad & \sum_{i=0}^{N-1}{\alpha E(P(j_i))+\beta t_f(P(j_i))}\\
  \textrm{s.t.} \quad & \forall j\in J \quad E(p(j))<E_{max}\\
    & \forall m\in M \quad c(m,t)\leq c(m)  \\
    & \forall b\in B \quad A(b,t)\geq 0,  \quad c(b,t)\leq c(b)\\
  \end{aligned}
  \end{equation}


\section{Main Results}
As the problem's dimension grows the optimization becomes computationally
infeasible. To address this, a subset of jobs within a specified
time window $T_w$ are scheduled at each segment of optimization. While the proposed solution may not
guarantee optimality, two heuristics are incorporated to enhance its
effectiveness.
\begin{enumerate}
  \item jobs are clustered, based on the closeness of their
  location, in $N_c$ clusters using K-means where $N_c$ is given by the user or is determined
  by a method like the Elbow method~\cite{cui2020introduction}. Then for each
  segment of optimization, $N_s$ jobs are selected from the clusters in proportion to the number of unassigned jobs in each cluster. Because the jobs are selected from different clusters the locations of selected jobs are different. Such a subset of jobs will be called \emph{diverse}. Let $select(N_s)$ select $N_s$ unassigned jobs in this manner.
  \item Optimization is divided into several segments, each in a time window, for optimizing
  the next segment a portion of the previous time window is considered again such
  that if it was possible to find a time slot for one of the next segment's jobs in the previous time window, the possibility can be used.
\end{enumerate}
A discrete-time model is assumed where time $t$, is represented as a natural number to
approximate continuous time, a time resolution parameter $\Delta
T$, which determines the granularity of the time intervals is used in the model. The time sequence of found solutions should be multiplied by $\Delta T$ to get real time.
The proposed algorithm consists of the following steps and is illustrated in
Fig~\ref{fig:overview}:

\begin{figure}[h]
  \centering
  \begin{tikzpicture}
    \pgfmathsetmacro{\heightBoxes}{0.7}
    \node[draw,    minimum width=1.5cm,    minimum height=\heightBoxes cm,    rounded corners=2mm] (cluster) at (0,0){Cluster};
    
    \node [draw,    rounded corners=2mm,    minimum width=1.5cm,    minimum height=\heightBoxes cm,    right=2cm of cluster]  (selector) {Selector};
    
    \node [draw,    rounded corners=2mm,     minimum width=1.5cm,     minimum height=\heightBoxes cm,   below= 0.3cm of selector]  (path) {Path Finder};
    
    \node [draw,    rounded corners=2mm,    minimum width=1.5cm,    minimum height=\heightBoxes cm,    below=0.3cm of path]  (Candidate) {Candidate Generator};
    
    \node [draw,    rounded corners=2mm,     minimum width=2cm,     minimum height=\heightBoxes cm,    below=0.3cm of Candidate] (solver) {Solver};

    \node [draw,    rounded corners=2mm,    minimum width=1.5cm,    minimum height=\heightBoxes cm,    below=1.315cm of cluster]  (time) {Time Window};

    \draw [->] (-2,0) --  (cluster.west)
        node[midway,above]{$jobs$};    
    
    \draw [->]  ($(cluster.north east)!0.33!(cluster.south east)$) -- ($(selector.north west)!0.33!(selector.south west)$);
    
    \draw [->]  (selector.south) -- (path.north);
    
    \draw [->]  (path.south) -- (Candidate.north);
    
    \draw [->]  (Candidate.south) -- (solver.north);
    
    \draw [->]  ($(time.north east)!0.33!(time.south east)$) -| ($(time.east)!0.2!(selector.west)$) |- (selector.west);
    
    \draw [->]  (time.east) -| ($(time.east)!0.5!(Candidate.west)$) |- (Candidate.west);
    
    \draw [->]  ($(time.north east)!0.66!(time.south east)$) -| ($(time.east)!0.2!(solver.west)$) |- ($(solver.north west)!0.33!(solver.south west)$);
    
    \draw [->] ($(solver.south west)!0.33!(solver.south east)$) -- ++(0,-0.25)  -|  ($(solver.west)-(0.5,0)$) node[left] at ($(solver.west)-(0.5,0.3)$) {\small Scheduled Jobs} --  (solver.west) ;
    
    \draw [->]  (solver.south) -- ++(0,-0.5) -| ($(time.west)-(0.5,0)$) -- (time.west) ;
  \end{tikzpicture}
  \caption{An overview of the proposed algorithm.}
  \label{fig:overview}
\end{figure}
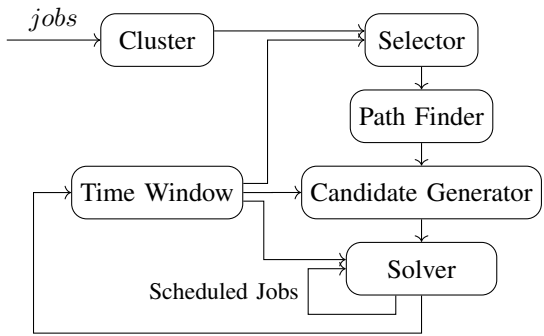

\begin{enumerate}
\item\label{item:enter base path}Find possible paths between base stations without passing through jobs or machines;
\item\label{item:select subset job} Select a diverse subset of jobs;
\item\label{item:find path} Find paths for selected jobs;
\item\label{item:time_window} Set the time window $T_w$.
\item\label{item:candidate} Append possible timing to paths for each job in the
  selected subset, to be called a candidate schedule.
\item Add constraints regarding capacity, processing time and availability of
  the machines used in each candidate schedule.
\item\label{item:solve boolean linear program} Solve the associate boolean linear program.
\end{enumerate}
Steps~\ref{item:select subset job} to~\ref{item:solve boolean linear program}
are repeated until all the jobs are scheduled.

Step~\ref{item:enter base path} is to exchange transporters between base
stations without performing any jobs. This is required when there is no
available transporter in the base stations within the feasible paths for a
specific job (i.e., a path with a length smaller than the transporter's range). In
such cases, a transporter can be transferred and recharged at an intermediary
base station.

In step~\ref{item:find path}, to comply with the limited range restriction, a predefined number of facility-disjoint shortest paths (i.e., paths that are at least different in one of the facilities assigned with minimum energy needed) for each job are found. Each path is a sequence $p=b_{D_k},j_i,m_{b,2},m_{c,3},\cdots,m_{z,S-2},b_{R_l}$ that has the same properties of the facility sequence of $P$ and must satisfy $E(p)<E_{max}$. An implementation of Dijkstra's single source shortest paths algorithm\footnote{https://pypi.org/project/Dijkstar/} is used to find the shortest path
and then a node of each found path is discarded to find a different shortest path without
the discarded nodes to make the paths facility-disjoint. A time window in step~\ref{item:time_window} is selected and a set of candidate
schedules are built in step~\ref{item:candidate} with Algorithm~\ref{alg:create candidates} for all the paths generated in step~\ref{item:find path}. Each
candidate $\rho_c(j_i)$ is like $P(j_i)$, with all its properties, where $\rho_{c_0}$ (i.e. the candidate's dispatching time) is
selected from the time window $T_w$ and $\rho_{c_1},\cdots, \rho_{c_{S-1}}$ are calculated based
on the transportation and processing times. Additionally, each candidate has a
boolean decision accessible by $value(\rho_c)$. Google's
OR-Tools\footnote{https://developers.google.com/optimization} is selected as the solver, and the
candidate value is added as an integer variable to the solver that can take only 0 or 1 for each candidate schedule. Value 1 specifies that the candidate will be used in the solution The proposed method is represented in Algorithm~\ref{alg:overal_optimizer} where it
fetchs, from the user: the jobs $J$, $N_s$, $N_c$, the moving horizon time ($\tau_h$) which specifies the time duration that is added to the time window at
each segment of optimization, and $\gamma$ to specify how much the current
segment's time window overlaps with the previous. The minimum number of
paths $P_{min}$ that should exist for a job (i.e., the minimum number of facility-disjoint paths to search for that are complying with the range limitation). Also, the maximum number of paths $P_{max}$ to skip
searching for more disjoint paths. Then, the time window is calculated. Let
$\tau$ be the finish time when all assigned jobs in the last optimization
segment are finished. The time window for the following segment of optimization
is determined in Algorithm~\ref{alg:overal_optimizer} as the interval
[$\tau-\gamma\tau_h,\tau+\tau_h$]. Then using $select(N_s)$ a set of jobs
for the following segment of optimization is selected. For each job in this set
Algorithm~\ref{alg:findPaths} finds at most $P_{max}$ paths for each job or
throws a warning if $P_{min}$ paths do not exist for a job. Consequently,
algorithm~\ref{alg:create candidates} is used to create candidate schedules
based on possible paths and the calculated time window. Finally, the constraints
are imposed and the optimization for each segment is done with
Algorithm~\ref{alg:milp}. This process is repeated until all the jobs are
scheduled.

Algorithm~\ref{alg:findPaths} uses Algorithm~\ref{alg:job_graph} to make a graph\_job for each job to find at least $P_{min}$ and at most $P_{max}$ candidate paths going through require machines and the base stations. This graph\_job is constructed with all the machines and only the selected job as its nodes. The base stations are duplicated as base\_dispatch and base\_recieve to distinguish when a transporter is dispatched from a base station and when a transporter returns. A virtual start node before base\_dispatch and a virtual end node after base\_recieve stage are added with zero time and energy edges to all bases, so that we can search for paths from the start node to the end node. Each edge has two properties: the transportation time and energy required to traverse that edge. Between each consequent stage, edges with the required transportation time and energy are created.
Then Algorithm~\ref{alg:findPaths} uses Dijkstar to find the shortest path, concerning the required energy and keeps it as a candidate path if it requires less energy than the maximum energy $E_{max}$ of a transporter. To find more facility-disjoint paths, nodes of a point in the discrete cartesian space of the set of founded paths (i.e., the space made of the nodes of each path as points on each axis) are eliminated from the job graph and then another shortest path is searched for. If less than $P_{min}$ paths are found for a job the algorithm returns a warning. This is to inform the user that there are not enough candidate paths. Particularly, if $P_{min}$ is set to 1 this means that a feasible path does not exist for the job and this job can not be assigned.
As there can be a feasible path for a job from a base station that does not contain a transporter, Algorithm~\ref{alg:findPaths} finds the possibility of sending a transporter from a base station to another base station without any job assigned as an intermediary base station.

To address the optimization task at each segment, Algorithm~\ref{alg:milp} is employed. Here the $add(,)$ function is to add candidate values to a constraint container. Let $\sigma=0$ be an empty constraint container then $add(\sigma,c1)$ and $add(\sigma,c2)$ will result in $\sigma=value(c1)+value(c2)$. The $sub(,)$ function is defined likewise with $-$ as the operator. 
The Constraints considered for the solver are:
\begin{itemize}
  \item Constraints for previously assigned candidates. That is for each candidate in the previous segment with value=1 a constraint is added to keep the value equal to 1 in the next segments.
  \item Constraints regarding jobs being assigned only once. Let $j_i$ be a job, all the candidates that are associated with $j_i$ (i.e., $\rho^1=j_i$) are added to the job's constraint container $\sigma[j_i]$ (i.e., $add(\sigma[j_i],\rho)$) and then for each job, a constraint is added to the solver such that the job's constraint container equals 1 (i.e., for each job $solver.addConstraint(\sigma[j_i]=1)$).
  \item Constraints regarding machines. Each machine has a processing time and a capacity which means that at any time there must be fewer ongoing operations than its capacity. Let $m_{i,s}$ be a machine at stage $s$, to comply with this restriction each candidate $\rho$ which $m_{i,s}\in \rho$ is added to the machine's constraint container $\sigma[m_{i,s}]$ for an interval with the duration of the processing time of the machine (i.e., $\forall t\in [\rho_s,\rho_s+pt(\rho^s)]$), $add(sigma[m_{i,s}],\rho)$.
  \item One of the constraints of the base stations is the conservation of transporters, which implies that the number of transporters dispatched from a base station can not be more than the number of transporters received plus the number of transporters initially existing in the base station. Let $b_i$ be a base station. To address this constraint any candidate that has $\rho^0=b_i$ (i.e., the transporter is dispatched from $b_i$) is subtracted from the availability constraint container of $b_i$ for an interval starting from $\rho$ to the end of the current time window (i.e., $\forall t\geq \rho_0 \cap T_w$,  $sub(\sigma_A[b_i][t],\rho)$). Also, a base station can be functioning like a machine (e.g., maintenance of transporters or integration of chargers in base stations). Thus, when a transporter is received, it gets ready to dispatch only after a processing time. In this case, any candidate that has $\rho^{S-1}=b_i$ is added to the availability constraint container of $b_i$ for an interval starting from the arriving time plus processing time of $b_i$ to the end of the time window (i.e., $\forall t\geq \rho_{S-1}+pt(b_i) \cap T_w$  $add(\sigma_A[b_i][t],\rho)$). Finally, the net number of all the dispatched and received transporters for any base station must be bigger or equal to the negative of initially existing transporters of $b_i$ that $\sigma_A[b_i]\geq -b_i(0)$.
  \item Another constraint related to the base stations is the capacity of the base station. Such that there is a maximum space and there can not be more than a certain number of transporters in a specific base station at any time. To respect this constraint any candidate that has $\rho^0=b_i$ is subtracted from the capacity constraint container of $b_i$ for an interval starting from $\rho_0$ to the end of the current time window (i.e., $\forall t\geq \rho_0 \cap T_w$  $sub(\sigma_c[b_i][t],\rho)$). When a transporter is received any candidate that has $\rho^{S-1}=b_i$ is added to the capacity constraint container of $b_i$ for an interval starting from the receiving time $\rho_{S-1}$ to the end of the time window (i.e., $\forall t\geq \rho_{S-1} \cap T_w$  $add(\sigma_c[b_i][t],\rho)$). Finally, the net number of all the dispatched and received transporters for any base station must be less or equal to the capacity of that base station $\sigma_c[b_i]\leq c(b_i)$.
  \end{itemize}

The objective function defined in~\eqref{eq:objective} is implemented as the summation of $(\alpha\ E(\rho) + \beta\ t_f(\rho)) value(\rho)$ of all candidate schedules. After the optimization is done in each segment, jobs with candidates whose value is assigned to be 1 are deleted from their cluster, the time window slides forward and the algorithm iterates until there are no more jobs to assign. If in a segment no job is assigned the algorithm throws a warning which means that the time window is not long enough or the job needs a transporter from a base station that is empty and there is no possibility of transferring a transporter to that base station from another base station.

\begin{algorithm}[h]
  \caption{Job's Graph Generator}\label{alg:job_graph}
  \Fn{$graph\_job(j)$}{
    $graph=Null$;\\
    \ForEach{$b_d \in Base\_dispatch$}{
      $graph.addEdge(start,b_d,0,0)$;\\  
      $graph.addEdge($
      $b_d,j,te(b_d,j),tt(b_d,j))$;\\
    }
    \ForEach{$stage$ \textbf{such that} $\exists stage.next()$}{
      \ForEach{$n \in stage$}{
        \ForEach{$m \in stage.next()$}{
        $graph.addEdge($
        $n,m,te(n,m)e,tt(n,m))$;\\         
        
      }  
    }
    }
    \ForEach{$b_r \in Base\_recieve$}{
      $graph.addEdge(b_r,end,0,0)$\;  
    }
    \Return$graph$
  }
\end{algorithm}

\begin{algorithm}[h]
  \caption{Possible Paths}\label{alg:findPaths}
  \Fn{$possible\_paths(jobs,E_{max},P_{min},$ $P_{max})$}{
  $paths=Null$;\\
  \ForEach{$job\in jobs$}{
    $graph\_j=graph\_job(job)$;
    $paths\_j=Null$;\\
    \Repeat{$size(paths\_j)\leq P_{max}$}{
      $p=finder(graph\_j,paths\_j,E_{max})$;\\
      \If{$p=Null$}{
        \textbf{warning}; 
        \textbf{break};\\
      }
      $paths\_j.append(p)$;\\
    }
    $paths.extend(path\_j)$;\\
  }
  \ForEach{$b \in base\_stage$}{
    \ForEach{$br \in base\_stage/base$}{
      $p=b,b_r$;\\
      \If{$E(p)\leq E_{max}$}{
        $paths.append(p)$;\\
    }
  }
  }
  \Return{$paths$}
  }
  \Fn{$finder(graph\_j,paths\_j,E_{max})$}{
    \ForEach{$point\in Cartesian(paths\_j)$}{
      $p=Dijkstar(graph\_j/point,start,end)$\;
      \If{$E(p)\leq E_{max}$ \& $p\neq Null$ }{
        \Return{$p$}\;
      }
    }
    \Return{$Null$}\;
  }
\end{algorithm}

\begin{algorithm}[h]
  \caption{Candidate Generator}\label{alg:create candidates}
  \Fn{$\rho\_gen(paths,T_w)$}{
    \ForEach{$p \in paths$}{
        \For{$t_d \in T_w$}{
          $\rho^0=p^0$; $\rho_0=t_d$;\\
          \ForEach{$1 \leq s<S$}{
            $\rho^{s}=p^s$;\\
            $\rho_{s}=\rho_{s-1}+tt(\rho^{s-1},p^{s})+pt(\rho^{s-1})attached(\rho^{s-1})$;\\
          }
          $candidates.append(\rho)$
        }
    }
    \Return$candidates$
  }

\end{algorithm}

\begin{algorithm}
  
  \caption{Solver}\label{alg:milp}
  \Fn{$solver(candidates,T_w,schedule=Null)$}{
    \ForEach{$sc\in schedule$ }{
      $solver.addConstraint(value(sc)==1)$;\\
    }
    \ForEach{$\rho \in candidates \cup schedule$}{
      \ForEach{$0\leq s <S$}{
        \If{$type(\rho^s)=job$}{
          $add(\sigma[\rho^s],\rho)$;\\
        }
        \ElseIf{$type(\rho^s)=machine$}{
          \For{$\rho_s \leq t \leq \rho_s+pt(\rho^s)$}{
            $\sigma[\rho^s][t].add(\rho)$;\\
          }
        }
        \ElseIf{$type(\rho^s)=base\_dispatch$}{
          \ForEach{$t>\rho_0 \cap T_w$}{
            $\sigma_c[\rho^s][t].sub(\rho)$;\\
            $\sigma_A[\rho^s][t].sub(\rho)$;\\
           }
        }
        \ElseIf{$type(\rho^s)=base\_recieve$}{
          \ForEach{$t>\rho_{S-1}+pt(\rho^s)\cap T_w$}{
            $\sigma_A[\rho^s][t].add(\rho)$;\\
          }
          \ForEach{$t>\rho_{S-1} \cap T_w$}{
            $\sigma_c[\rho^s][t].add(\rho)$;\\
          }
        }
      }
      $solver.{addVar}(\rho)$\\
    }

      \ForEach{$j \in jobs$}{
        $solver.addConstraint(\sigma[j]=1)$
      }
      \ForEach{$m \in machines$}{
        \ForEach{$t \in \sigma[m].keys()$}{
          $solver.addConstraint(\sigma[m][t]\leq c(m)$\\
        }
        
      }
      \ForEach{$b \in bases$}{
        \ForEach{$t \in \sigma_c[b].keys()$}{
          $solver.addConstraint(\sigma_c[b]\leq c(b)$\\
        }
        \ForEach{$t \in \sigma_A[b].keys()$}{
          $solver.addConstraint(\sigma_A[b]\geq -b(0))$\\
        }
      }
      \ForEach{$\rho \in candidates$}{
        $solver.addObjective((\alpha\ E(\rho) + \beta t_f(\rho))\rho.value)$
      }
      $solution=solver.minimize()$\\
      \If{$\exists\ sl$\text{$\in$} $solution$ \textbf{such that} $value(sl)=1$}{
        \ForEach{$sl$ \text{$\in$} $solution$ \textbf{such that} $value(sl)=1$}{
          $schedule.append(sl)$;\\
          $makespan=\max(makespan,sl_{S-1})$;\\
          $sl.cluster.delete(sl^1)$
        }
        \Else{
          Warning
        }
      }
      
      \Return{$schedule,\ makespan$}
  }
  
\end{algorithm}




\begin{algorithm}
  \caption{Optimizer}\label{alg:overal_optimizer}
  \Fn{$Optimizer(jobs,N_c,\tau_h,\gamma=1,makespan=0,N_s,E_{max},P_{min}=1,P_{max}=\infty)$}{
  $scheduled=Null$;
  $cluster(jobs,property,N_c)$;\\
  \Repeat{$All\ jobs\ assigned$}{
    $start\_time=\min(makespan-\tau_h\gamma,0)$;\\
    $end\_time=makespan+\tau_h$;\\
    $T_w=[start\_time,end\_time]$;\\
    $selected\_jobs=select(N_s)$;\\
    $possible\_path=possible\_paths(selected\_jobs,E_{max},P_{min},P_{max})$;\\
    $candidates=\rho\_gen(possible\_path,T_w)$;\\
    $scheduled, makespan=solve(candidates,T_w,scheduled)$;\\
  }
  }
\end{algorithm}

It is worth mentioning that the proposed algorithm is flexible to several extensions as modifying the path-finding part and the objective function can be simply applied. As examples, consider the following remarks:
\begin{remark}
  For jobs that need to skip a stage, say stage $s$, it is possible to change $stage.next()$ to $stage.next().next()$ when creating the job's graph in Algorithm~\ref{alg:job_graph} at stage $s-1$. Therefore, candidate paths will not use machines in stage $s$.
\end{remark}
\begin{remark}
  The objective function~\eqref{eq:objective} tends to minimize the average time of finishing all the jobs. If the time for accomplishing the jobs that take more time is of concern, it is possible to change the objective function in Algorithm~\ref{alg:milp} to $(\alpha\ E(\rho) + \beta\ t_f(\rho)^k) value(\rho)$, where $k$ can be any natural number. The higher the $k$ is chosen, the job that finishes last will be more costly. Therefore, the finish time in each segment will be minimized.
  Please note that this objective function is still linear concerning $value(\rho)$.
\end{remark}
\begin{remark}
 If a job has a predefined time of accomplishment $t_{ac}$, the job can be given more priority in $select(N_s)$ and in a time window whose end time is later than the assigned time of accomplishment. Then, $\zeta(\rho_1-t_{ac})^2 value(\rho)$ could be added to the objective function to penalize the non-punctuality of accomplishing a job.   
\end{remark}

\section{Simulation Results}
\begin{figure}[ht]
  \centering
  \includegraphics[width=\columnwidth]{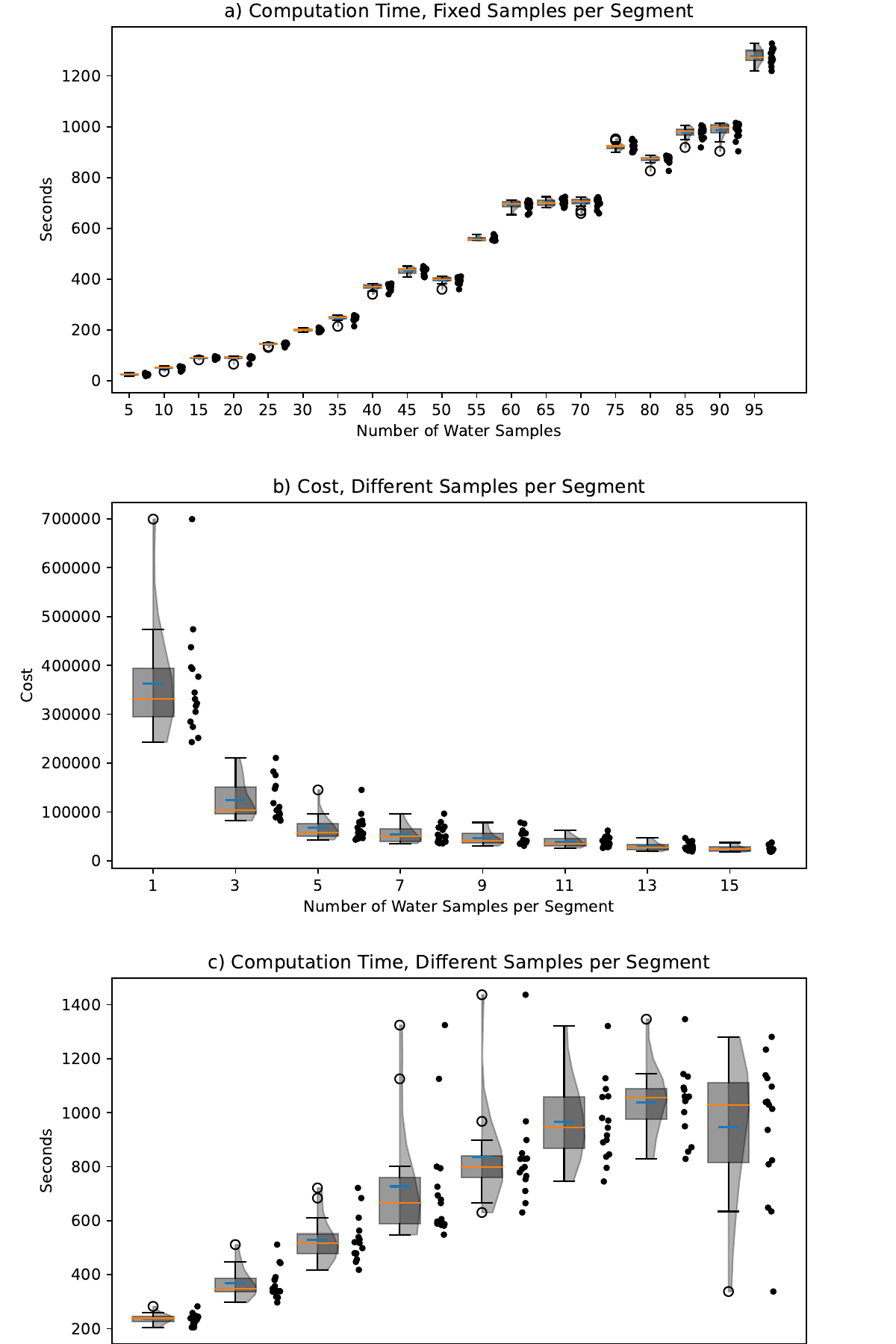}
  \caption{15 experiments: a) The computation time with fixed 5 jobs per segment and different numbers of jobs. b) The cost with different jobs per segment choices and a fixed 100 total jobs. c) The computation time with different choices of jobs per segment and 100 jobs}
  \label{fig:all}
\end{figure}

As an experiment, the proposed method is implemented in an automated system for water sampling along the coast of Nova Scotia using a swarm of flying robots that could be considered one of the current large-scale water sampling programs in which shellfish production is regulated by the Canadian Shellfish Sanitation Program (CSSP) and requires determination of water sanitation levels on shellfish beds by monitoring the presence of E. Coli. This section shows simulation results on 15 randomly generated experiments with 100 jobs, 10 drones, and 10 laboratories. The code and parameters are publicly accessible~\footnote{\url{https://git.mistlab.ca/kkhateri/mrs_paper}}. Figure~\ref{fig:all}.a, depicts the computation time of the proposed method as a function of the number of water samples (jobs). The complexity of the method increases linearly with the number of samples while the number of samples per segment of optimization remains constant. This demonstrates the scalability of the proposed method in handling a large number of samples. In Fig~\ref{fig:all}.b, the cost difference resulting from different choices of samples per segment of optimization is shown. Please note that in the case of one sample per segment, the method is equivalent to the greedy algorithm and as the number of samples per segment is increased it approaches optimality exponentially such that even with a small choice of samples per segment the cost is reduced by a large margin while the computation time is still in orders of seconds. The computation time as shown in Fig~\ref{fig:all}.c increases with the number of samples per segment. This highlights the trade-off between the optimality of the solution and the computation time. Users can adjust this trade-off based on their specific requirements and available computation resources.

\section{Conclusions}
We have shown a method to divide the joint machine-transporter scheduling for
multistage jobs into sub-problems, which provides a significant improvement in
terms of solution quality over a greedy algorithm. In addition, our method has
parameters to adjust the trade-off between computation time and solution
quality. We show that our method's computation time grows linearly with the
number of tasks, ensuring scalability.




\bibliographystyle{IEEEtran}
\bibliography{main.bib}

\end{document}